\documentclass[11pt,a4paper]{article}
\usepackage[hyperref]{acl2017}
\usepackage{times}
\usepackage{latexsym}
\usepackage{amsmath}
\usepackage{amssymb}
\usepackage{bbm}
\usepackage{graphicx}
\usepackage{ifthen}
\usepackage[utf8]{inputenc}
\usepackage{color,colortbl}
\usepackage{url}
\usepackage{booktabs}
\usepackage{subcaption}
\usepackage{multirow}
\usepackage{array}

\newif\ifcomment

\aclfinalcopy 

\newcommand\p[1]{\ensuremath{\left( #1 \right)}} 
\newcommand\pb[1]{\ensuremath{\left[ #1 \right]}} 
\newcommand\pc[1]{\ensuremath{\left\{ #1 \right\}}} 

\newcommand\refsec[1]{Section~\ref{sec:#1}}

\newcommand\reffig[1]{Figure~\ref{fig:#1}}

\newcommand\reftab[1]{Table~\ref{tab:#1}}
\newcommand\refapp[1]{Appendix~\ref{sec:#1}}

\ifthenelse{\isundefined{\definition}}{\newtheorem{definition}{Definition}}{}
\ifthenelse{\isundefined{\assumption}}{\newtheorem{assumption}{Assumption}}{}
\ifthenelse{\isundefined{\hypothesis}}{\newtheorem{hypothesis}{Hypothesis}}{}
\ifthenelse{\isundefined{\proposition}}{\newtheorem{proposition}{Proposition}}{}
\ifthenelse{\isundefined{\theorem}}{\newtheorem{theorem}{Theorem}}{}
\ifthenelse{\isundefined{\lemma}}{\newtheorem{lemma}{Lemma}}{}
\ifthenelse{\isundefined{\corollary}}{\newtheorem{corollary}{Corollary}}{}
\ifthenelse{\isundefined{\alg}}{\newtheorem{alg}{Algorithm}}{}
\ifthenelse{\isundefined{\example}}{\newtheorem{example}{Example}}{}

\ifcomment

\newcommand\hh[1]{\textcolor{blue}{[HH: #1]}}
\newcommand\me[1]{\textcolor{green}{[ME: #1]}}
\newcommand\ab[1]{\textcolor{orange}{[AB: #1]}}
\else

\newcommand\hh[1]{}
\newcommand\me[1]{}
\newcommand\ab[1]{}
\fi

\newcommand{\kba}{$\text{KB}_\text{A}$}
\newcommand{\kbb}{$\text{KB}_\text{B}$}
\newcommand{\ent}[1]{{\small\texttt{#1}}}
\newcommand{\MF}{{MutualFriends}}
\newcommand{\utterance}[1]{``#1''}
\newcommand{\cond}{\ensuremath{\,|\,}}

\newcommand{\dkg}{{DynoNet}}
\newcommand{\skg}{{StanoNet}}
\newcommand{\rl}{{Rule}}

\newcommand{\fl}{{\small Flnt}}
\newcommand{\cor}{{\small Crct}}
\newcommand{\co}{{\small Coop}}

\newcommand{\hu}{{\small Human}}

\newcommand{\act}[1]{{\small{\textsf {#1}}}}

\newcolumntype{L}[1]{>{\raggedright\let\newline\\\arraybackslash\hspace{0pt}}m{#1}}
\newcolumntype{C}[1]{>{\centering\let\newline\\\arraybackslash\hspace{0pt}}m{#1}}
\newcolumntype{R}[1]{>{\raggedleft\let\newline\\\arraybackslash\hspace{0pt}}m{#1}}

\newcommand{\ul}[1]{\underline{#1}}

\title{Learning Symmetric Collaborative Dialogue Agents with Dynamic Knowledge Graph Embeddings}

\author{
    He He \and Anusha Balakrishnan \and Mihail Eric \and Percy Liang \\
    Computer Science Department, Stanford University \\
    {\tt \{hehe,anusha28,meric,pliang\}@cs.stanford.edu}
}

\date{}

\begin{document}

\maketitle

\begin{abstract}

We study a \emph{symmetric collaborative dialogue} setting
in which two agents, each with private knowledge,
must strategically communicate to achieve a common goal.
The open-ended dialogue state in this setting poses new challenges for existing dialogue systems.
We collected a dataset of 11K human-human dialogues,
which exhibits interesting lexical, semantic, and strategic elements.
To model
both structured knowledge and unstructured language,
we propose a neural model with dynamic knowledge graph embeddings
that evolve as the dialogue progresses.
Automatic and human evaluations show that our model is both more effective
at achieving the goal and more human-like than baseline neural and rule-based models.

\end{abstract}

\section{Introduction}
\label{sec:intro}

Current task-oriented dialogue
systems~\cite{young2013pomdp,wen2017network,dhingra2017information}
require a pre-defined dialogue state (e.g., slots such as food type and price range
for a restaurant searching task)
and a fixed set of dialogue acts (e.g., request, inform).
However, human conversation often requires richer dialogue states and more
nuanced, pragmatic dialogue acts.
Recent open-domain chat systems
~\citep{shang2015neural,serban2015building,sordoni2015neural,li2016persona,lowe2017ubuntu,mei2017coherent}
learn a mapping directly from previous utterances to the next utterance.
While these models capture open-ended aspects of dialogue,
the lack of structured dialogue state prevents them from
being directly applied to settings
that require interfacing with structured knowledge.

\begin{figure}[ht]
  {\footnotesize
  Friends of agent A: \vspace{2mm}\\
  \begin{tabular}{llll}
    \toprule
    Name     & School   & Major              & Company  \\ \midrule
    Jessica  & Columbia & Computer Science   & Google   \\
    Josh     & Columbia & Linguistics        & Google   \\
    ...      & ...      & ...                & ...      \\ \bottomrule
  \end{tabular}

    \newcommand\pA{A}
    \newcommand\pB{B}
    \medskip
  \begin{tabular}{@{}l@{}}
    \pA: Hi! Most of my friends work for Google \\
    \pB: do you have anyone who went to columbia? \\
    \pA: \emph{Hello?}  \\
    \pA: I have Jessica a friend of mine  \\
    \pA: and Josh, both went to columbia \\
    \pB: \emph{or anyone working at apple?}  \\
    \pB: SELECT (Jessica, Columbia, Computer Science, Google) \\
    \pA: SELECT (Jessica, Columbia, Computer Science, Google)
  \end{tabular}
  }
  \caption{\label{fig:example_dialog}
    An example dialogue from the \MF{} task
    in which two agents, A and B,
    each given a private list of a friends,
    try to identify their mutual friend.
    Our objective is to build an agent that can perform the task with a human.
    Cross-talk (\refsec{data_stat}) is \emph{italicized}.
  }
\end{figure}

In order to bridge the gap between the two types of systems,
we focus on a \emph{symmetric collaborative dialogue} setting,
which is task-oriented but encourages open-ended dialogue acts.
In our setting, two agents, each with a private list of items with attributes,
must communicate to identify the unique shared item.
Consider the dialogue in \reffig{example_dialog},
in which two people are trying to find their mutual friend.
By asking \utterance{do you have anyone who went to columbia?},
B is suggesting that she has some Columbia friends, and that they probably work at Google.
Such conversational implicature is lost when interpreting the utterance as
simply an information request.
In addition, it is hard to define a structured state that captures the diverse semantics in many utterances
(e.g., defining ``most of'', ``might be''; see details in \reftab{type_example}).

To model both structured and open-ended context,
we propose the \emph{\underline{Dy}namic K\underline{no}wledge Graph \underline{Net}work} (\dkg{}),
in which the dialogue state is modeled as a knowledge graph with an embedding for each node (\refsec{overview}).
Our model is similar to EntNet~\cite{henaff2017tracking}
in that node/entity embeddings are updated recurrently given new utterances.
The difference is that we structure entities as a knowledge graph;
as the dialogue proceeds,
new nodes are added and new context is propagated on the graph.
An attention-based mechanism~\cite{bahdanau2015neural} over the node embeddings drives generation of new utterances.
Our model's use of knowledge graphs
captures the grounding capability of classic task-oriented systems
and the graph embedding provides the representational flexibility of neural models.

The naturalness of communication in the symmetric collaborative setting
enables large-scale data collection:
We were able to crowdsource around 11K human-human dialogues on Amazon Mechanical Turk (AMT) in less than 15 hours.\footnote{The dataset is available publicly at \url{https://stanfordnlp.github.io/cocoa/}.}
We show that the new dataset calls for more flexible representations beyond fully-structured states (\refsec{data}).

In addition to conducting the third-party human evaluation adopted by most work~\cite{liu2016evaluate,li2016diversity,li2016rl},
we also conduct partner evaluation~\cite{wen2017network} where AMT workers rate their conversational partners
(other workers or our models) based on
fluency, correctness, cooperation, and human-likeness.
We compare \dkg{} with baseline neural models and a strong rule-based system.
The results show that \dkg{} can perform the task with humans efficiently and naturally;
it also captures some strategic aspects
of human-human dialogues.

The contributions of this work are:
(i) a new symmetric collaborative dialogue setting and a large dialogue corpus that pushes the boundaries of existing dialogue systems;
(ii) \dkg{}, which integrates semantically rich utterances with structured knowledge to represent open-ended dialogue states;
(iii) multiple automatic metrics based on bot-bot chat and a comparison of third-party and partner evaluation.

\section{Symmetric Collaborative Dialogue}
\label{sec:problem}
We begin by introducing a collaborative task between two agents
and describe the human-human dialogue collection process.
We show that our data exhibits diverse, interesting language phenomena.

\subsection{Task Definition}
In the symmetric collaborative dialogue setting,
there are two agents, A and B, each with
a private knowledge base---\kba{} and \kbb{}, respectively.
Each knowledge base includes a list of \emph{items},
where each item has a value for each \emph{attribute}.
For example, in the \MF{} setting,
\reffig{example_dialog},
items are friends and attributes are name, school, etc.
There is a shared item that A and B both have;
their goal is to converse with each other to determine the shared item
and select it.
Formally, an agent is a mapping from its private KB and the dialogue thus far (sequence of
utterances) to the next utterance to generate or a selection. A dialogue is considered \textit{successful} when both agents correctly select the shared item.
This setting has parallels in human-computer collaboration where each agent has complementary expertise.

\begin{table*}[ht]
\centering
{\footnotesize
\setlength\tabcolsep{0.5ex}
\begin{tabular}{C{1cm}C{0.8cm}L{6.5cm}L{7cm}}
\toprule
Type & \% & Easy example & Hard example \\
\midrule
Inform & 30.4 &
I know a \ul{judy}. / I have someone who \ul{studied the bible} in the \ul{afternoon}. &
\textbf{About equal} \ul{indoor} and \ul{outdoor} friends / {\bf me too}. his major is \ul{forestry} / \textbf{might be} \ul{kelly}\\
Ask & 17.7 &
Do any of them like \ul{Poi}? / What does your \ul{henry} do? &
What can you tell me about our friend? / \textbf{Or maybe } \ul{north park college}?\\
Answer & 7.4 &
None of mine did / Yup / They do. / Same here. &
yes 3 of them / No he likes \ul{poi} / yes if \ul{boston college} \\
\bottomrule
\end{tabular}
}
\caption{\label{tab:type_example}
Main utterance types and examples.
We show both standard utterances whose meaning can be represented by simple logical forms (e.g., \act{ask}(\ent{indoor})),
and open-ended ones which require more complex logical forms (difficult parts in bold).
Text spans corresponding to entities are underlined.
}
\end{table*}

\begin{table*}[ht]
\centering
{\footnotesize
\setlength\tabcolsep{0ex}
\begin{tabular}{C{2.5cm}L{13.5cm}}
\toprule
Phenomenon & Example \\
\midrule
Coreference    & (I know one \ul{Debra}) does she like the \ul{indoors}? / (I have two friends named \ul{TIffany}) at \ul{World airways}? \\
Coordination   & keep on going with the \ul{fashion} / Ok. let's try something else. / go by hobby / great. select him. thanks!\\
Chit-chat      & Yes, that is {\bf good ole} \ul{Terry}. / All \ul{indoorsers}! {\bf my friends hate nature}\\
Categorization & same, most of mine are \bf{female} too / Does any of them {\bf names start with B}\\
Correction     & I know one friend into \ul{Embroidery} - her name is \ul{Emily}. {\bf Sorry -- \ul{Embroidery} friend is named \ul{Michelle}} \\
\bottomrule
\end{tabular}
}
\caption{\label{tab:phenomenon_example}
Communication phenomena in the dataset.
Evident parts is in bold and
text spans corresponding to an entity are underlined.
For coreference, the antecedent is in parentheses.
}
\end{table*}

\subsection{Data collection}
\label{sec:data}

We created a schema with 7 attributes and approximately 3K entities (attribute values).
To elicit linguistic and strategic variants,
we generate a random scenario for each task by varying
the number of items (5 to 12),
the number attributes (3 or 4),
and the distribution of values for each attribute (skewed to uniform).
See \refapp{schema} and \ref{sec:scenario} for details of schema and scenario generation.

\begin{figure}[ht]
\centering
\includegraphics[width=0.5\textwidth]{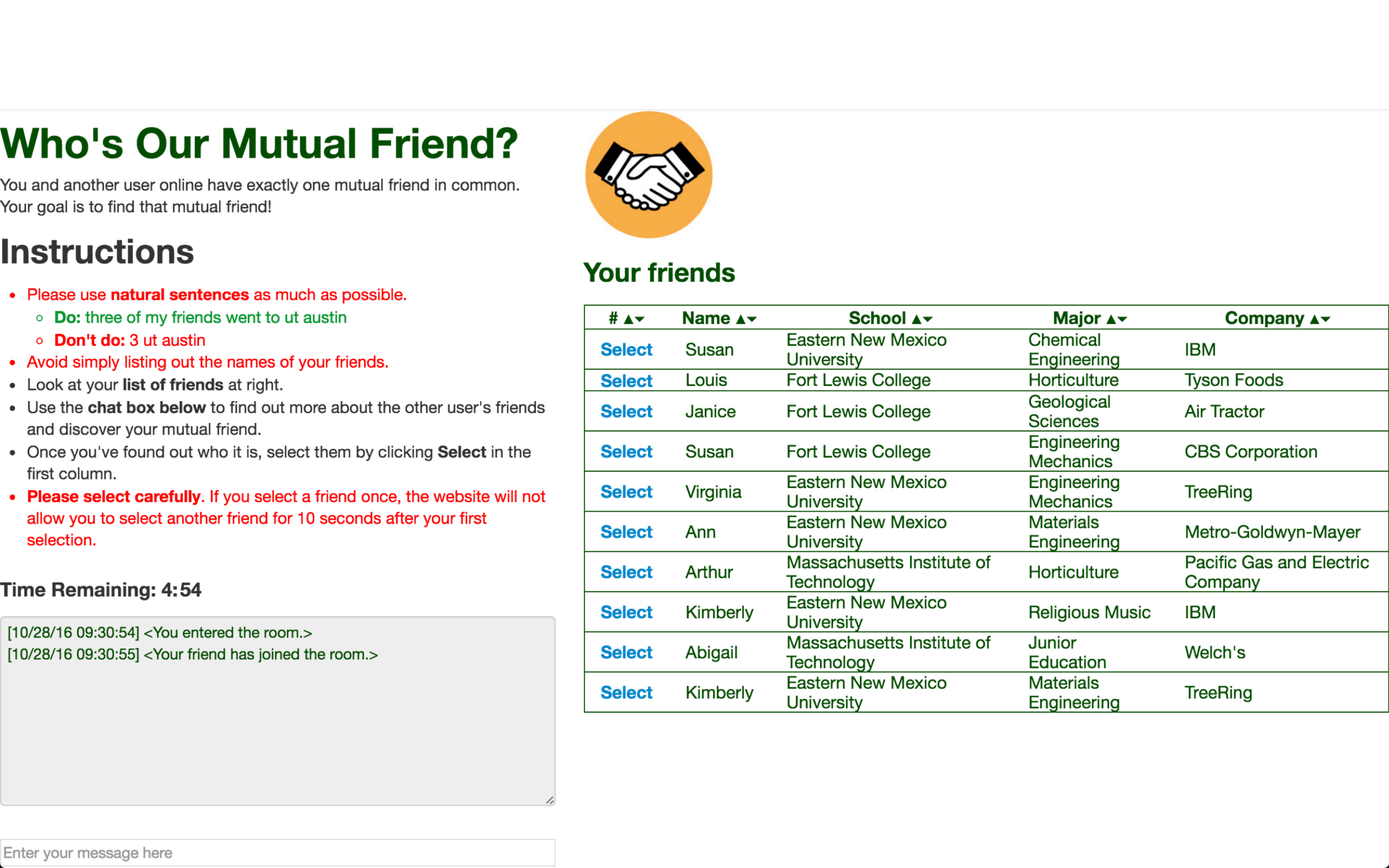}
\caption{Screenshot of the chat interface.}\label{fig:website}
\end{figure}

We crowdsourced dialogues on AMT by randomly pairing up workers
to perform the task within 5 minutes.\footnote{If the workers exceed the time limit, the dialogue is marked as unsuccessful (but still logged).}
Our chat interface is shown in \reffig{website}.
To discourage random guessing, we prevent workers from selecting more than once every 10 seconds.
Our task was very popular and
we collected 11K dialogues over a period of 13.5 hours.\footnote{Tasks are put up in batches; the total time excludes intervals between batches.}
Of these, over 9K dialogues are successful. Unsuccessful dialogues are usually the result of either worker leaving the chat prematurely.

\subsection{Dataset statistics}
\label{sec:data_stat}
We show the basic statistics of our dataset in \reftab{gen-statistics}.
An utterance is defined as a message sent by one of the agents.
The average utterance length is short due to the informality of the chat, however,
an agent usually sends multiple utterances in one turn.
Some example dialogues are shown in \reftab{human-bot-chats} and \refapp{human-bot-chats}.

\begin{table}[ht]
\centering
{\footnotesize
\begin{tabular}{lr}
\toprule
\# dialogues & 11157 \\
\# completed dialogues & 9041 \\
Vocabulary size & 5325 \\
Average \# of utterances & 11.41 \\
Average time taken per task (sec.) & 91.18 \\
Average utterance length (tokens) & 5.08 \\
Number of linguistic templates\footnotemark  & 41561 \\
\bottomrule
\end{tabular}
}
 \caption{Statistics of the \MF{} dataset.}\label{tab:gen-statistics}
\end{table}
\footnotetext{Entity names are replaced by their entity types.}

We categorize utterances into coarse types---\act{inform}, \act{ask}, \act{answer}, \act{greeting}, \act{apology}---by pattern matching (\refapp{utterance_type}).
There are 7.4\% multi-type utterances, and 30.9\% utterances contain more than one entity. 
In \reftab{type_example}, we show example utterances with rich semantics that cannot be sufficiently represented by traditional slot-values.
Some of the standard ones are also non-trivial due to coreference and logical compositionality.

Our dataset also exhibits some interesting communication phenomena.
Coreference occurs frequently when people check multiple attributes of one item.
Sometimes mentions are dropped, as an utterance simply continues from the partner's utterance.
People occasionally use external knowledge to group items with out-of-schema attributes (e.g., gender based on names, location based on schools).
We summarize these phenomena in \reftab{phenomenon_example}.
In addition, we find 30\% utterances involve cross-talk where the conversation does not progress linearly (e.g., italic utterances in \reffig{example_dialog}),
a common characteristic of online chat~\citep{ivanovic2005dialogue}. 

One strategic aspect of this task is choosing the order of attributes to mention.
We find that people tend to start from attributes with fewer unique values,
e.g., \utterance{all my friends like morning} given the \kbb{} in \reftab{human-bot-chats},
as intuitively it would help exclude items quickly given fewer values to check.\footnote{Our goal is to model human behavior thus we do not discuss the optimal strategy here.}
We provide a more detailed analysis of strategy in \refsec{eval} and \refapp{strategy}.

\section{Dynamic Knowledge Graph Network}
\label{sec:overview}
\begin{figure*}[ht]
\centering
\includegraphics[width=\textwidth]{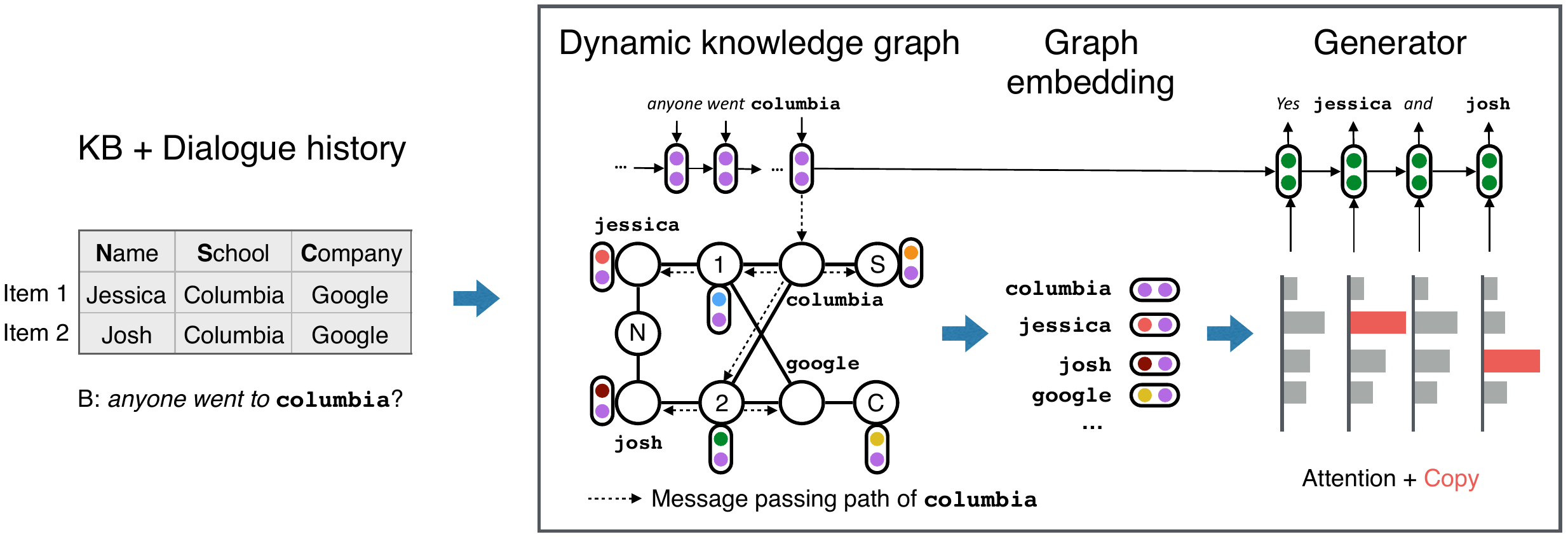}
\caption{\label{fig:overview}
Overview of our approach.
First, the KB and dialogue history (entities in {\bf\ent{bold}}) is mapped to a graph.
Here, an item node is labeled by the item ID and
an attribute node is labeled by the attribute's first letter.
Next, each node is embedded using relevant utterance embeddings through message passing.
Finally, an LSTM generates the next utterance based on attention over the node embeddings.
}
\end{figure*}
The diverse semantics in our data motivates us to combine
unstructured representation of the dialogue history  
with structured knowledge.
Our model consists of three components shown in \reffig{overview}:
(i) a dynamic knowledge graph,
which represents the agent's private KB and shared dialogue history as a graph (\refsec{kg}),
(ii) a graph embedding over the nodes (\refsec{graph_embed}),
and (iii) an utterance generator (\refsec{gen}).

The knowledge graph represents entities and relations in the agent's private KB,
e.g., \ent{item-1}'s \ent{company} is \ent{google}.
As the conversation unfolds, utterances are embedded and incorporated into node embeddings of mentioned entities.
For instance, in \reffig{overview}, \utterance{anyone went to columbia} updates the embedding of \ent{columbia}.
Next, each node recursively passes its embedding to neighboring nodes 
so that related entities (e.g., those in the same row or column) also receive information from the most recent utterance.
In our example, \ent{jessica} and \ent{josh} both receive new context when \ent{columbia} is mentioned.
Finally, the utterance generator, an LSTM, produces the next utterance by attending to
the node embeddings. 

\subsection{Knowledge Graph}
\label{sec:kg}
Given a dialogue of $T$ utterances,
we construct graphs $(G_t)_{t=1}^T$ over the KB and dialogue history for agent A.\footnote{
It is important to differentiate perspectives of the two agents as they have different KBs.
Thereafter we assume the perspective of agent A,
i.e., accessing \kba{} for A only,
and refer to B as the partner.
}
There are three types of nodes:
item nodes,
attribute nodes,
and entity nodes.
Edges between nodes represent their relations.
For example, 
\ent{(item-1, hasSchool, columbia)} means that the first item has attribute \ent{school} whose value is \ent{columbia}.
An example graph is shown in \reffig{overview}.
The graph $G_t$ is updated based on utterance $t$
by taking $G_{t-1}$ and adding a new node for any entity mentioned in utterance $t$ but not in \kba{}.\footnote{
We use a rule-based lexicon to link text spans to entities.
See details in \refapp{lexicon}.
}

\subsection{Graph Embedding}
\label{sec:graph_embed}
Given a knowledge graph,
we are interested in computing a vector representation for each node $v$ that captures both its unstructured context from the dialogue history and its structured context in the KB.
A node embedding $V_t(v)$ for each node $v \in G_t$ is built from three parts:
structural properties of an entity defined by the KB,
embeddings of utterances in the dialogue history,
and message passing between neighboring nodes.

\paragraph{Node Features.}
Simple structural properties of the KB often govern what is talked about;
e.g., a high-frequency entity is usually interesting to mention
(consider \utterance{All my friends like dancing.}).
We represent this type of information as a feature vector $F_t(v)$,
which includes the degree and type (item, attribute, or entity type) of node $v$,
and whether it has been mentioned in the current turn.
Each feature is encoded as a one-hot vector and they are concatenated to form $F_t(v)$.

\paragraph{Mention Vectors.}
A mention vector $M_t(v)$ contains unstructured context from utterances relevant to node $v$ up to turn $t$. 
To compute it,
we first define the utterance representation $\tilde{u}_t$
and the set of relevant entities $E_t$.
Let $u_t$ be the embedding of utterance $t$ (\refsec{gen}).
To differentiate between the agent's and the partner's utterances,
we represent it as $\tilde{u}_t=\pb{u_t \cdot \mathbbm{1}_{\pc{u_t \in U_{\text{self}}}}, u_t \cdot \mathbbm{1}_{\pc{u_t \in U_{\text{partner}}}}}$,
where $U_{\text{self}}$ and $U_{\text{partner}}$ denote sets of utterances generated by the agent and the partner,
and $[\cdot,\cdot]$ denotes concatenation.
Let $E_t$ be the set of entity nodes mentioned in utterance $t$ if utterance $t$ mentions some entities,
or utterance $t-1$ otherwise.\footnote{
Relying on utterance $t-1$ is useful when utterance $t$ answers a question, e.g.,
\utterance{do you have any google friends?} \utterance{No.}
}
The mention vector $M_t(v)$ of node $v$ incorporates the current utterance
if $v$ is mentioned and inherits $M_{t-1}(v)$ if not:
\begin{align}
M_t(v) &= \lambda_t M_{t-1}(v) + (1 - \lambda_t) \tilde{u}_t; \\
\lambda_t &=
\begin{cases}
  \sigma\p{W^\text{inc} \pb{M_{t-1}(v), \tilde{u}_t}} & \text{if $v \in E_t$}, \\
1 & \text{otherwise}. \nonumber
\end{cases}
\end{align}
Here, $\sigma$ is the sigmoid function and
$W^\text{inc}$ is a parameter matrix.

\paragraph{Recursive Node Embeddings.}
We propagate information between nodes according to the structure of the knowledge graph.
In \reffig{overview}, given \utterance{anyone went to columbia?},
the agent should focus on her friends who went to Columbia University.
Therefore, we want this utterance to be sent to item nodes connected to \ent{columbia},
and one step further to other attributes of these items because
they might be mentioned next as relevant information, e.g.,
\ent{jessica} and \ent{josh}.

We compute the node embeddings recursively, analogous to belief propagation:
\begin{align}
V_t^k(v) =& \max_{v' \in N_t(v)} \tanh \\
  &\p{W^\text{mp} \pb{V_t^{k-1}(v'), R(e_{v\rightarrow v'})}}, \nonumber
\end{align}
where $V_t^k(v)$ is the depth-$k$ node embedding at turn $t$ and
$N_t(v)$ denotes the set of nodes adjacent to $v$.
The message from a neighboring node $v'$ depends on its embedding at depth-$(k-1)$, 
the edge label $e_{v\rightarrow v'}$
(embedded by a relation embedding function $R$),
and a parameter matrix $W^\text{mp}$.
Messages from all neighbors are aggregated by
$\max$, the element-wise max operation.\footnote{Using sum or mean slightly hurts performance.}
Example message passing paths are shown in \reffig{overview}.

The final node embedding is the concatenation of embeddings at each depth:
\begin{align}
V_t(v) = \pb{V_t^0(v), \ldots, V_t^K(v)},
\label{eqn:node_embedding}
\end{align}
where $K$ is a hyperparameter (we experiment with $K \in \{0,1,2\}$) and
$V_t^0(v) = \pb{F_t(v), M_t(v)}$.

\subsection{Utterance Embedding and Generation}
\label{sec:gen}

We embed and generate utterances
using Long Short Term Memory (LSTM) networks
that take the graph embeddings into account.

\paragraph{Embedding.}

On turn $t$,
upon receiving an utterance consisting of $n_t$ tokens,
$x_t = (x_{t,1}, \dots, x_{t,n_t})$,
the LSTM maps it to a vector as follows:
\begin{align}
  h_{t,j} = \text{LSTM}_{\text{enc}}(h_{t,j-1}, A_t(x_{t,j})),
\end{align}
where $h_{t,0} = h_{t-1,n_{t-1}}$, and
$A_t$ is an \emph{entity abstraction} function, explained below.
The final hidden state $h_{t,n_t}$ is used as the utterance embedding $u_t$, which updates the mention vectors
as described in \refsec{graph_embed}.

In our dialogue task, the identity of an entity is unimportant.
For example, replacing \ent{google} with \ent{alphabet} in \reffig{example_dialog}
should make little difference to the conversation.
The role of an entity is determined instead by its relation to other entities
and relevant utterances.
Therefore, we define the abstraction $A_t(y)$ for a word $y$ as follows:
if $y$ is linked to an entity $v$, then
we represent an entity by its type (\ent{school}, \ent{company} etc.) embedding
concatenated with its current node embedding: $A_t(y) = [E_{\text{type}(y)}, V_t(v)]$.
Note that $V_t(v)$ is determined only by its structural features and its context.
If $y$ is a non-entity, then $A_t(y)$ is the word embedding of $y$ concatenated with a zero vector of the same dimensionality as $V_t(v)$.
This way, the representation of an entity only depends on its structural properties given by the KB
and the dialogue context,
which enables the model to generalize to unseen entities at test time.

\paragraph{Generation.}

Now, assuming we have embedded utterance $x_{t-1}$ into $h_{t-1,n_{t-1}}$ as described above,
we use another LSTM to generate utterance $x_t$.
Formally, 
we carry over the last utterance embedding $h_{t,0} = h_{t-1,n_{t-1}}$ and define:
\begin{align}
  h_{t,j} = \text{LSTM}_{\text{dec}}(h_{t,j-1}, \pb{A_t(x_{t,j}), c_{t,j}}),
\end{align}
where $c_{t,j}$ is a weighted sum of node embeddings in the current turn:
$c_{t,j} = \sum_{v \in G_t} \alpha_{t,j,v} V_t(v)$,
where $\alpha_{t,j,v}$ are the attention weights over the nodes.
Intuitively, high weight should be given to relevant entity nodes as shown in \reffig{overview}.
We compute the weights through standard attention mechanism \citep{bahdanau2015neural}:
\begin{align*}
\alpha_{t,j} &= \text{softmax}(s_{t,j}),\\
s_{t,j,v} &= w^\text{attn} \cdot \tanh\p{W^\text{attn} \pb{h_{t,j-1}, V_t(v)}},
\end{align*}
where vector $w^\text{attn}$ and $W^\text{attn}$ are parameters.

Finally, we define a distribution over both words in the vocabulary and nodes in $G_t$
using the copying mechanism of \newcite{jia2016recombination}:
\begin{align*}
p(x_{t,j+1} &= y \cond G_t, x_{t,\le j}) \propto \exp\p{W^{\text{vocab}} h_{t,j} + b}, \\
p(x_{t,j+1} &= r(v) \cond G_t, x_{t,\le j}) \propto \exp\p{s_{t,j,v}},
\end{align*}
where
$y$ is a word in the vocabulary,
$W^{\text{vocab}}$ and $b$ are parameters,
and $r(v)$ is the realization of the entity represented by node $v$,
e.g., \ent{google} is realized to ``Google'' during copying.\footnote{
We realize an entity by sampling from the empirical distribution of its surface forms found in the training data.
}

\section{Experiments}
\label{sec:experiments}

We compare our model with a rule-based system and a baseline neural model.
Both automatic and human evaluations are conducted to test the models in terms of fluency, correctness, cooperation, and human-likeness.
The results show that \dkg{} is able to converse with humans in a coherent and strategic way.

\subsection{Setup}

We randomly split the data into train, dev, and test sets (8:1:1).
We use a one-layer LSTM with 100 hidden units
and 100-dimensional word vectors for both the encoder and the decoder (\refsec{gen}).
Each successful dialogue is turned into two examples, each from the perspective of one of the two agents.
We maximize the log-likelihood of all utterances in the dialogues.
The parameters are optimized by AdaGrad~\cite{duchi10adagrad} with an initial learning rate of 0.5.
We trained for at least 10 epochs; after that, training stops if there is no improvement on the dev set for 5 epochs.
By default, we perform $K=2$ iterations of message passing to compute node embeddings (\refsec{graph_embed}).
For decoding, we sequentially sample from the output distribution with a softmax temperature of 0.5.\footnote{
Since selection is a common `utterance' in our dataset and
neural generation models are susceptible to over-generating common sentences,
we halve its probability during sampling.
}
Hyperparameters are tuned on the dev set.

\ab{Maybe add a reminder of what \dkg{} is (the acronym was introduced and used only once in the intro section) and explicitly show how \skg{} differes from it.}
We compare \dkg{} with  
its static cousion (\skg{}) and
a rule-based system (\rl{}). 
\skg{} uses $G_0$ throughout the dialogue,
thus the dialogue history is completely contained in the LSTM states
instead of being injected into the knowledge graph.
\rl{} maintains weights for each entity and each item in the KB to decide
what to talk about and which item to select.
It has a pattern-matching semantic parser, a rule-based policy, and a templated generator.
See \refapp{rule} for details.

\subsection{Evaluation}
\label{sec:eval}
We test our systems in two interactive settings:
bot-bot chat
and bot-human chat.
We perform both automatic evaluation and human evaluation.

\begin{table*}[ht]
\centering
{\footnotesize
\setlength{\tabcolsep}{.55em}
\begin{tabular}{c|c|cc|ccc|ccccc|cccc}
\toprule
System & $\ell\downarrow$
& $L_u$ & $H$ 
& $C\uparrow$ & $C_T\uparrow$ & $C_S\uparrow$ 
& Sel & Inf & Ask & Ans & Greet 
& \#Ent$_1$ & $|\text{Attr}_1|$ & \#Ent & \#Attr \\
\midrule
Human       & -           
& 5.10  & 4.57  
& .82 & .07 & .38 
& .21 & .31 & .17 & .08 & .08 
&.55 & .35 & 6.1 & 2.6\\
\midrule
\rl{}  & -           
& 7.61  & 3.37  
& .90 & .05 & {\bf.29} 
& .18 & {\bf.34} & .23 & .00 & .12
& .24 & .61 & 9.9 & 3.0 \\
\skg{}   & 2.20        
& 4.01  & {\bf4.05}  
& .78 & .04 & .18 
& .19 & .26 & .12 & .23 & {\bf.09} 
& .61 & {\bf .19} & 7.1 & 2.9\\
\dkg{}  & {\bf 2.13}  
& 3.37  & 3.90  
& {\bf .96} & {\bf.06} & .25 
& {\bf.22} & .26 & {\bf.13} & .20 & .12 
& {\bf .55} & .18 & {\bf5.2} & {\bf2.5}\\
\bottomrule
\end{tabular}
}
\caption{\label{tab:auto-eval} Automatic evaluation on human-human and bot-bot chats on test scenarios.
We use $\uparrow$ / $\downarrow$ to indicate that higher / lower values are better;
otherwise the objective is to match humans' statistics.
Best results (except Human) are in bold.
Neural models generate shorter (lower $L_u$) but more diverse (higher $H$) utterances.
Overall, their distributions of utterance types match those of the humans'.
(We only show the most frequent speech acts therefore the numbers do not sum to 1.)
\rl{} is effective in completing the task (higher $C_S$), but
it is not information-efficient given the large number of attributes (\#Attr) and entities (\#Ent) mentioned.
}
\end{table*}

\paragraph{Automatic Evaluation.}
First, we compute the cross-entropy ($\ell$) of a model on test data.
As shown in \reftab{auto-eval}, \dkg{} has the lowest test loss.
Next, we have a model chat with itself on the scenarios from the test set.\footnote{
    We limit the number of turns in bot-bot chat to be the maximum number of turns humans took in the test set (46 turns).}
We evaluate the chats with respect to language variation, effectiveness, and strategy.

For language variation, we report the average utterance length $L_u$ and the unigram entropy $H$ in \reftab{auto-eval}.
Compared to \rl{},
the neural models tend to generate shorter utterances~\cite{li2016diversity,serban2017hierarchical}.
However, they are more diverse;
for example, questions are asked in multiple ways such as
\utterance{Do you have ...},
\utterance{Any friends like ...},
\utterance{What about ...}.

At the discourse level, we expect the distribution of a bot's utterance types to match the distribution of human's.
We show percentages of each utterance type in \reftab{auto-eval}.
For \rl{}, the decision about which action to take is written in the rules,
while \skg{} and \dkg{} learned to behave in a more human-like way,
frequently informing and asking questions.

To measure effectiveness, we compute the overall success rate ($C$)
and the success rate per turn ($C_T$) and per selection ($C_S$).
As shown in \reftab{auto-eval}, humans are the best at this game,
followed by \rl{} which is comparable to \dkg{}.

Next, we investigate the strategies leading to these results. 
An agent needs to decide which entity/attribute to check first
to quickly reduce the search space.
We hypothesize that humans tend to first focus on a majority entity and an attribute with fewer unique values (\refsec{data_stat}).
For example, in the scenario in \reftab{human-bot-chats},
\ent{time} and \ent{location} are likely to be mentioned first.
We show the average frequency of first-mentioned entities (\#Ent$_1$)
and the average number of unique values for first-mentioned attributes ($|\text{Attr}_1|$)
in \reftab{auto-eval}.\footnote{
Both numbers are normalized to $[0,1]$ with respect to all entities/attributes in the corresponding KB.}
Both \dkg{} and \skg{} successfully match human's starting strategy
by favoring entities of higher frequency and attributes of smaller domain size. 

To examine the overall strategy, we show the average number of attributes (\#Attr) and entities (\#Ent) mentioned during the conversation in \reftab{auto-eval}.
Humans and \dkg{} strategically focus on a few attributes and entities,
whereas \rl{} needs almost twice entities to achieve similar success rates.
This suggests that the effectiveness of \rl{} mainly comes from large amounts of unselective information,
which is consistent with comments from their human partners.

\begin{table*}[ht]
\centering
{\footnotesize
\begin{tabular}{c|ccc|llll|llll}
\toprule
\multirow{2}{*}{System} & 
\multirow{2}{*}{$C$} & \multirow{2}{*}{$C_T$} & \multirow{2}{*}{$C_S$} & 
\multicolumn{4}{c|}{Partner eval} & \multicolumn{4}{c}{Third-party eval} \\
& & & & \fl{} & \cor{} & \co{} & \hu{} & \fl{} & \cor{} & \co{}  & \hu{}  \\
\midrule
Human & 
.89 & .07 & .36 &
4.2$^{rds}$ & 4.3$^{rds}$ & 4.2$^{rds}$ & 4.1$^{rds}$ &
4.0& 4.3$^{ds}$ & 4.0$^{ds}$ & 4.1$^{rds}$ \\
\midrule
\rl{} & 
{\bf.88} & {\bf .06} & {\bf .29} &
3.6 & 4.0 & 3.5 & 3.5 &
4.0& {\bf 4.4}$^{hds}$ & {\bf 3.9}$^{s}$ & {\bf 4.0}$^{s}$\\
\skg{}& 
.76 & .04 & .23 &
3.5 & 3.8 & 3.4 & 3.3 &
4.0& 4.0& 3.8 & 3.8\\
\dkg{}& 
.87 & .05 & .27 &
{\bf3.8}$^{s}$ & 4.0 & {\bf3.8}$^{rs}$ & {\bf3.6}$^s$ &
4.0& 4.1& 3.9 & 3.9\\
\bottomrule
\end{tabular}
}
\caption{\label{tab:human-eval} Results on human-bot/human chats.
Best results (except Human) in each column are in bold.
We report the average ratings of each system.
For third-party evaluation, we first take mean of each question then average the ratings.
\dkg{} has the best partner satisfaction in terms of fluency (\fl{}), correctness (\cor{}), cooperation (\co{}), human likeness (\hu{}).
  The superscript of a result indicates that its advantage over other systems ($r$: \rl{}, $s$: \skg{}, $d$: \dkg{}) is statistically significant with $p<0.05$ given by paired $t$-tests.
}
\end{table*}

\newcommand{\eos}{$||$}
\renewcommand{\bot}[1]{{\bf #1}}
\definecolor{LightCyan}{rgb}{0.88,1,1}
\begin{table*}[ht]
{\footnotesize
\setlength\tabcolsep{0.5ex}
\begin{tabular}{lllll}
\multicolumn{5}{l}{Friends of A}\\
\toprule
ID & Name & Company & Time & Location \\
\midrule
1 & Kathy     &TRT Holdings             &afternoon     &indoor  \\
2 & Jason     &Dollar General           &afternoon     &indoor\\
3 & Johnny    &TRT Holdings             &afternoon     &outdoor\\
4 & Frank     &SFN Group                &afternoon     &indoor\\
5 & Catherine &Dollar General           &afternoon     &indoor\\
6 & Catherine &Weis Markets             &afternoon     &indoor\\
\rowcolor{LightCyan}
7 & Kathleen  &TRT Holdings             &morning       &indoor\\
8 & Lori      &TRT Holdings             &afternoon     &indoor\\
9 & Frank     &L\&L Hawaiian Barbecue   &afternoon     &outdoor\\
\bottomrule
\end{tabular}
\quad
\begin{tabular}{lllll}
\multicolumn{5}{l}{Friends of B}\\
\toprule
ID & Name & Company & Time & Location \\
\midrule
1 & Justin     &New Era Tickets                &morning     &indoor\\
\rowcolor{LightCyan}
2 & Kathleen   &TRT Holdings                   &morning     &indoor\\
3 & Gloria     &L\&L Hawaiian Barbecue         &morning     &indoor\\
4 & Kathleen   &Advance Auto Parts             &morning     &outdoor\\
5 & Justin     &Arctic Cat                     &morning     &indoor\\
6 & Anna       &Dollar General                 &morning     &indoor\\
7 & Steven     &SFN Group                      &morning     &indoor\\
8 & Wayne      &R.J. Corman Railroad Group     &morning     &indoor\\
9 & Alexander  &R.J. Corman Railroad Group     &morning     &indoor\\
\bottomrule
\end{tabular}
\setlength\tabcolsep{0.5ex}
\begin{tabular}{L{7.5cm}L{7.5cm}}
\toprule
A: Human\quad B: Human & A: \dkg{} \quad B: Human\\
\midrule
\begin{tabular}[t]{L{7cm}}
A: Hi \\
B: hey \eos{} i have one outdoor\\
A: I have 4 TRT Holdings \eos{} I have 2 outdoor one Johnny, other Frank\\
B: i only have one TRT Holdings - Kathleen\\
A: SELECT 7 \quad B: SELECT 2 \\
\end{tabular} &
\begin{tabular}[t]{l}
A: \bot{hi}\\
B: Hello \eos{} all my friends prefer morning\\
A: \bot{1 of my morning likes the outdoors}\\
B: and all like indoor except for one\\
A: \bot{do they work for trt holdings?}   \\
B: Kathleen?\\
A: \bot{SELECT 7}  \quad B: SELECT 2 \\
\end{tabular} \\
\midrule
A: \skg{} \quad B: Human & A: Human \quad B: \rl{}\\
\midrule
\begin{tabular}[t]{l}
A: \bot{Hello}     \\
B: hi\\
A: \bot{Hello \eos{} I have one morning person.}     \\
B: all of my friends like mornings\\
A: \bot{My friend prefers afternoon works at trt holdings.}     \\
B: what is their name?\\
A: \bot{Likes indoors.}     \\
B: what is your fiend who likes morning name?\\
A: \bot{They work for trt holdings.}     \\
B: SELECT 2  \quad A: \bot{SELECT 7} \\
\end{tabular} &
\begin{tabular}[t]{l}
B: \bot{hiya}\\
A: hEY\\
B: \bot{I have 1 indoors and kathleen.}\\
A: Most of mine are indoors.\\
B: \bot{SELECT 1}\\
A: I have one morning and rest afternoon.\\
B: \bot{Do you have any friend working at l hawaiian?}\\
A: I don't know Justin\\
B: \bot{I have 1 alexander.} \\
...\\
\end{tabular} \\
\bottomrule
\end{tabular}
} 
\caption{\label{tab:human-bot-chats}
Examples of human-bot chats. The mutual friend is highlighted in blue in each KB.
Bots' utterances are in bold and
selected items are represented by item IDs.
Only the first half of the human-\rl{} chat is shown due to limited space. 
Multiple utterances of one agent rae separated by $||$.
}
\end{table*}

\paragraph{Partner Evaluation.}
We generated 200 new scenarios
and put up the bots on AMT using the same chat interface that was used for data collection.
The bots follow simple turn-taking rules explained in \refapp{turn-taking}.
Each AMT worker is randomly paired with
\rl{}, \skg{}, \dkg{}, or another human (but the worker doesn't know which),
and we make sure that all four types of agents are tested in each scenario at least once.
At the end of each dialogue, humans are asked to rate their partner in terms of
fluency, correctness, cooperation, and human-likeness from 1 (very bad) to 5
(very good), along with optional comments.

We show the average ratings (with significance tests) in \reftab{human-eval} and the histograms in \refapp{histograms}.
In terms of fluency, the models have similar performance since the utterances are usually short.
Judgment on correctness is a mere guess since the evaluator cannot see the partner's KB;
we will analyze correctness more meaningfully in the third-party evaluation below.

Noticeably, \dkg{} is more cooperative than the other models.
As shown in the example dialogues in \reftab{human-bot-chats},
\dkg{} cooperates smoothly with the human partner,
e.g., replying with relevant information about morning/indoor friends 
when the partner mentioned that all her friends prefer morning and most like indoor.
\skg{} starts well
but doesn't follow up on the morning friend,
presumably because the \ent{morning} node is not updated dynamically 
when mentioned by the partner.
\rl{} follows the partner poorly. 
In the comments, the biggest complaint about \rl{} was that it was not `listening' or `understanding'.
Overall, \dkg{} achieves better partner satisfaction, especially in cooperation.

\paragraph{Third-party Evaluation.}
We also created a \emph{third-party evaluation} task,
where an independent AMT worker is shown a conversation and the KB of one of the agents;
she is asked to rate the same aspects of the agent as in the partner evaluation and provide justifications.
Each agent in a dialogue is rated by at least 5 people.

The average ratings and histograms are shown in \reftab{human-eval} and \refapp{histograms}.
For correctness, we see that \rl{} has the best performance since it always tells the truth,
whereas humans can make mistakes due to carelessness and the neural models can generate false information.
For example, in \reftab{human-bot-chats}, \dkg{} `lied' when saying that it has a morning friend who likes outdoor. 

Surprisingly, there is a discrepancy between the two evaluation modes in terms of cooperation and human-likeness.
Manual analysis of the comments indicates that
third-party evaluators focus less on the dialogue strategy
and more on linguistic features, probably because they were not fully engaged in the dialogue.
For example, justification for cooperation often mentions frequent questions and timely answers,
less attention is paid to what is asked about though.

For human-likeness, partner evaluation is largely correlated with coherence (e.g., not repeating or ignoring past information) and task success,
whereas third-party evaluators often rely on informality 
(e.g., usage of colloquia like ``hiya'',
capitalization, and abbreviation)
or intuition.
Interestingly, third-party evaluators noted most phenomena listed in \reftab{phenomenon_example}
as indicators of human-beings,
e.g., correcting oneself,
making chit-chat other than simply finishing the task.
See example comments in \refapp{comments}.

\subsection{Ablation Studies}
\label{sec:ablation}
Our model has two novel designs: 
entity abstraction and message passing for node embeddings.
\reftab{ablation} shows what happens if we ablate these.
When the number of message passing iterations, $K$, is reduced from 2 to 0,
the loss consistently increases.
Removing entity abstraction---meaning
adding entity embeddings to node embeddings and the LSTM input embeddings---also
degrades performance.
This shows that \dkg{} benefits from contextually-defined, structural node embeddings
rather than ones based on a classic lookup table.

\begin{table}[ht]
\centering
{\footnotesize
\begin{tabular}{lc}
\toprule
Model & $\ell$ \\
\midrule
\dkg{} (K = 2) & {\bf 2.16} \\
\dkg{} (K = 1) & 2.20 \\
\dkg{} (K = 0) & 2.26 \\
\dkg{} (K = 2) w/o entity abstraction & 2.21 \\
\bottomrule
\end{tabular}
}
\caption{
  \label{tab:ablation} Ablations of our model on the dev set
  show the importance of entity abstraction and message passing ($K=2$).
  }
\end{table}

\section{Discussion and Related Work}
\label{sec:discussion}
There has been a recent surge of interest in end-to-end task-oriented dialogue systems,
though progress has been limited by the size of available datasets~\cite{serban2015survey}.
Most work focuses on information-querying tasks,
using Wizard-of-Oz data collection~\cite{williams2016dstc,maluuba2016frames} or simulators~\cite{bordes2017learning,li2016user},
In contrast, collaborative dialogues are easy to collect as natural human conversations,
and are also challenging enough given the large number of scenarios and diverse conversation phenomena.
There are some interesting strategic dialogue datasets---settlers of Catan~\cite{afantenos2012developing} (2K turns)
and the cards corpus~\cite{potts2012cards} (1.3K dialogues),
as well as work on dialogue strategies~\cite{keizer2017negotiation,vogel2013emergence}, 
though no full dialogue system has been built for these datasets.

Most task-oriented dialogue systems follow the POMDP-based approach~\cite{williams2007partially,young2013pomdp}.
Despite their success~\cite{wen2017network,dhingra2017information,su2016continuous}, the requirement for handcrafted slots limits their scalability to new domains
and burdens data collection with extra state labeling.
To go past this limit, \newcite{bordes2017learning} proposed a Memory-Networks-based approach without domain-specific features.
However, the memory is unstructured and interfacing with KBs relies on API calls,
whereas our model embeds both the dialogue history and the KB structurally.
\newcite{williams2017dialog} use an LSTM to automatically infer the dialogue state,
but as they focus on dialogue control rather than the full problem,
the response is modeled as a templated action,
which restricts the generation of richer utterances.
Our network architecture is most similar to EntNet~\cite{henaff2017tracking},
where 
memories are also updated by input sentences recurrently.
The main difference is that our model allows information to be propagated between structured entities,
which is shown to be crucial in our setting (\refsec{ablation}).  

Our work is also related to language generation conditioned on knowledge bases~\cite{mei2016what,kiddon2016globally}.
One challenge here is to avoid generating false or contradicting statements,
which is currently a weakness of neural models. 
Our model is mostly accurate when generating facts
and answering existence questions about a single entity,
but will need a more advanced attention mechanism for generating utterances involving multiple entities,
e.g., attending to items or attributes first, then selecting entities;
generating high-level concepts before composing them to natural tokens~\cite{serban2017multiresolution}.

In conclusion, we believe the symmetric collaborative dialogue setting and our dataset
provide unique opportunities at the interface of traditional task-oriented dialogue and open-domain chat.
We also offered \dkg{} as a promising means for open-ended dialogue state representation.
Our dataset facilitates the study of pragmatics and human strategies in dialogue---a good stepping stone towards learning more complex dialogues such as negotiation.

\paragraph{Acknowledgments.}
This work is supported by DARPA Communicating with Computers (CwC)
program under ARO prime contract no. W911NF-15-1-0462. 
Mike Kayser worked on an early version of the project while he was at Stanford.
We also thank members of the Stanford NLP group for insightful discussions.

\paragraph{Reproducibility.}
All code, data, and experiments for this paper are
available on the CodaLab platform:
{\footnotesize \url{https://worksheets.codalab.org/worksheets/0xc757f29f5c794e5eb7bfa8ca9c945573}}.

\bibliographystyle{acl_natbib}
\bibliography{all}

\clearpage
\appendix
\section{Knowledge Base Schema}
\label{sec:schema}
The attribute set $\mathcal{A}$ for the MutualFriends task contains name, school, major, company, hobby, time-of-day preference, and location preference.
Each attribute $a$ has a set of possible values (entities) $\mathcal{E}_a$.
For name, school, major, company, and hobby, we collected a large set of values from various online sources.\footnote{Names: \url{https://www.ssa.gov/oact/babynames/decades/century.html}\\Schools: \url{http://doors.stanford.edu/~sr/universities.html}\\Majors: \url{http://www.a2zcolleges.com/majors}\\Companies: \url{https://en.wikipedia.org/wiki/List_of_companies_of_the_United_States}\\Hobbies: \url{https://en.wikipedia.org/wiki/List_of_hobbies}}
We used three possible values (morning, afternoon, and evening) for the time-of-day preference, and two possible values (indoors and outdoors) for the location preference. 

\section{Scenario Generation}
\label{sec:scenario}
We generate scenarios randomly to vary task complexity and elicit linguistic and strategic variants. 
A scenario $S$ is characterized by the number of items ($N_S$),
the attribute set ($\mathcal{A}_S$) whose size is $M_S$,
and the values for each attribute $a \in \mathcal{A}_S$ in the two KBs.

A scenario is generated as follows.
\begin{enumerate}
\item Sample $N_S$ and $M_S$
  uniformly from $\{5,\dots, 12\}$ and $\{3,4\}$ respectively.
\item Generate $\mathcal{A}_S$ by sampling $M_S$ attributes without replacement from $\mathcal{A}$.
\item For each attribute $a \in \mathcal{A}_S$, sample the concentration parameter $\alpha_a$ uniformly from the set $\{0.3,1,3\}$.
\item Generate two KBs by sampling
$N_S$ values for each attribute $a$ from a Dirichlet-multinomial distribution over the value set $\mathcal{E}_a$ with the concentration parameter $\alpha_a$.
\end{enumerate}
We repeat the last step until the two KBs have one unique common item.

\section{Chat Interface}
\label{sec:website}
In order to collect real-time dialogue between humans, we set up a web server and
redirect AMT workers to our website.
Visitors are randomly paired up as they arrive.
For each pair, we choose a random scenario, and randomly assign a KB to each dialogue participant.
We instruct people to play intelligently, to refrain from brute-force tactics (e.g., mentioning every attribute value), and to use grammatical sentences.
To discourage random guessing, we prevent users from selecting a friend (item) more than once every 10 seconds.
Each worker was paid \$0.35 for
a successful dialogue within a 5-minute time limit. 
We log each utterance in the dialogue along with timing information.

\section{Entity Linking and Realization}
\label{sec:lexicon}
We use a rule-based lexicon to link text spans to entities. For every entity in the schema, we compute different variations of its canonical name, including acronyms, strings with a certain edit distance, prefixes, and morphological variants. 
Given a text span, a set of candidate entities is returned by string matching.
A heuristic ranker then scores each candidate (e.g., considering whether the span is a substring of a candidate, the edit distance between the span and a candidate etc.). 
The highest-scoring candidate is returned.

A linked entity is considered as a single token and its surface form is ignored in all models.
At generation time, we realize an entity by sampling from the empirical distribution of its surface forms in the training set.

\section{Utterance Categorization}
\label{sec:utterance_type}
We categorize utterances into \act{inform}, \act{ask}, \act{answer}, \act{greeting}, \act{apology} heuristically by pattern matching.
\begin{itemize}
\item An \act{ask} utterance asks for information regarding the partner's KB.
We detect these utterances by checking for the presence of a `?' and/or a question word like ``do", ``does", ``what", etc. 
\item 
An \act{inform} utterance provides information about the agent's KB. We define it as an utterances that mentions entities in the KB and is not an \act{ask} utterance. 
\item An \act{answer} utterance simply provides a positive/negative response to a question, containing words like ``yes", ``no", ``nope", etc. 
\item A \act{greeting} utterance contains words like ``hi" or ``hello"; it often occurs at the beginning of a dialogue.
\item An \act{apology} utterance contains the word ``sorry", which is typically associated with corrections and wrong selections. 
\end{itemize}
See \reftab{phenomenon_example} and \reftab{type_example} for examples of these utterance types.

\section{Strategy}
\label{sec:strategy}
During scenario generation, we varied the number of attributes, the number of items in each KB,
and the distribution of values for each attribute.
We find that as the number of items and/or attributes grows, the dialogue length and the completion time also increase,
indicating that the task becomes harder.
We also anticipated that varying the value of $\alpha$ would impact the overall strategy
(for example, the order in which attributes are mentioned)
since $\alpha$ controls the skewness of the distribution of values for an attribute. 

\begin{figure}[ht]
\centering
\includegraphics[width=0.8\columnwidth]{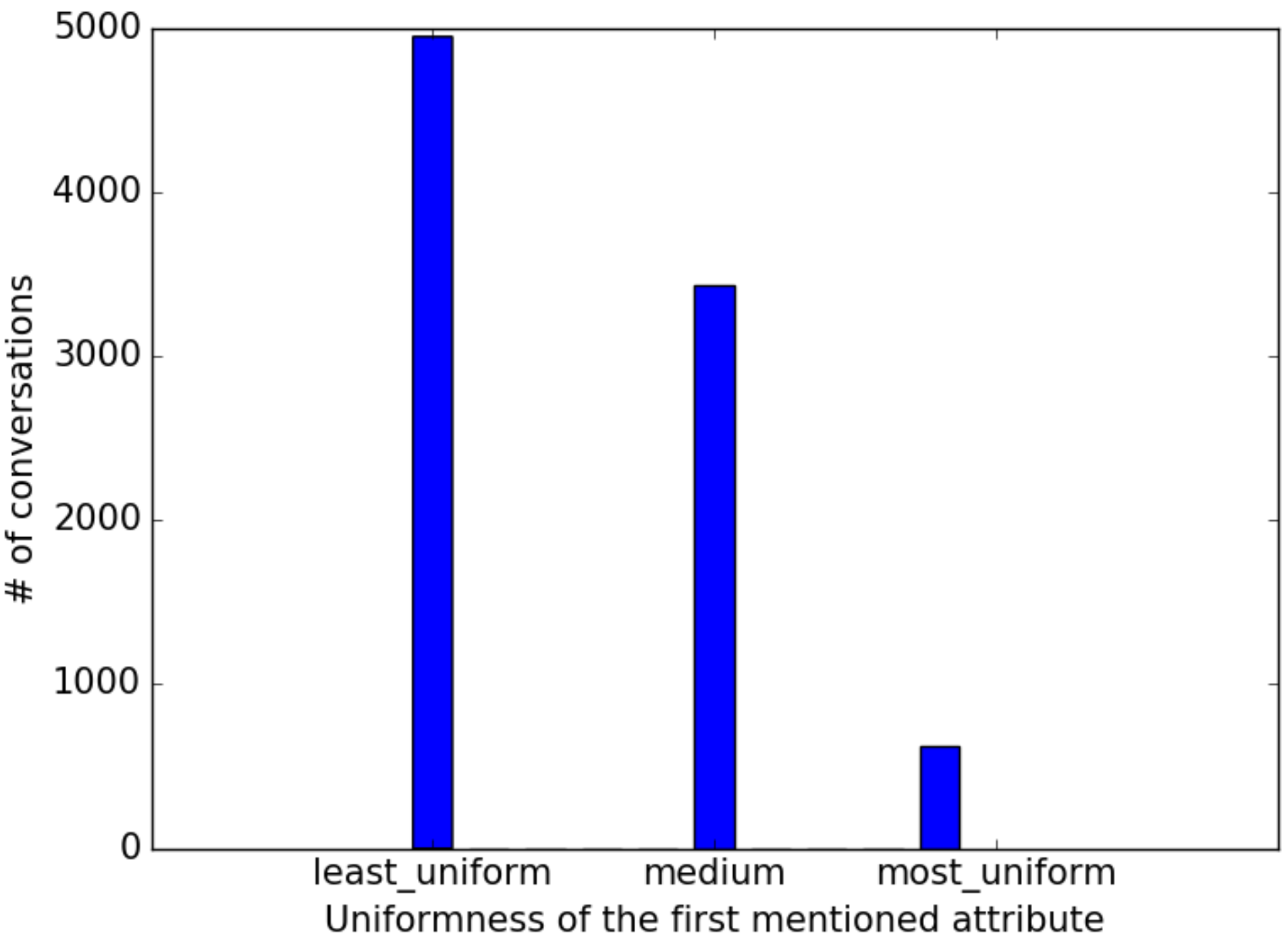}
\caption{Histogram of the first attribute mentioned in a dialogue. People tend to first mention attributes from very skewed (non-uniform) distributions. }\label{fig:attr_stats}
\end{figure}

\newcommand{\grp}[1]{{\small{\textsf {#1}}}}
On examining the data, we find that humans tend to first mention attributes with a more skewed (i.e., less uniform) distribution of values.
Specifically, we rank the $\alpha$ values of all attributes in a scenario (see step 3 in \refsec{scenario}),
and bin them into 3 distribution groups---\grp{least\_uniform}, \grp{medium}, and \grp{most\_uniform}, according to the ranking
where higher $\alpha$ values corresponds to more uniform distributions.\footnote{
    For scenarios with 3 attributes, each group contains one attributes.
    For scenarios with 4 attributes, we put the two attributes with rankings in the middle to \grp{medium}. }
In \reffig{attr_stats}, we plot the histogram of the distribution group of the first-mentioned attribute in a dialogues,
which shows that skewed attributes are mentioned much more frequently.

\section{Rule-based System}
\label{sec:rule}
The rule-based bot takes the following actions: 
greeting,
informing or asking about a set of entities,
answering a question,
and selecting an item.
The set of entities to inform/ask is sampled randomly given the entity weights.
Initially, each entity is weighted by its count in the KB.
We then increment or decrement weights of entities mentioned by the partner
and its related entities (in the same row or column),
depending on whether the mention is positive or negative.
A negative mention contains words like ``no'', ``none'', ``n't'' etc.
Similarly, each item has an initial weight of 1,
which is updated depending on the partner's mention of its attributes.

If there exists an item with weight larger than 1,
the bot selects the highest-weighted item with probability 0.3.
If a question is received, the bot informs facts of the entities being asked,
e.g., ``anyone went to columbia?'', ``I have 2 friends who went to columbia''.
Otherwise, the bot samples an entity set and randomly chooses between informing and asking about the entities.

All utterances are generated by sentence templates,
and parsing of the partner's utterance is done by entity linking and pattern matching (\refsec{utterance_type}).

\section{Turn-taking Rules}
\label{sec:turn-taking}
Turn-taking is universal in human conversations
and the bot needs to decide when to `talk' (send an utterance).
To prevent the bot from generating utterances continuously and forming a monologue,
we allow it to send at most one utterance if the utterance contains any entity, and two utterances otherwise.
When sending more than one utterance in a turn, the bot must wait for 1 to 2 seconds in between.
In addition, after an utterance is generated by the model (almost instantly),
the bot must hold on for some time to simulate message typing before sending.
We used a typing speed of 7 chars / sec and added an additional random delay between 0 to 1.5s after `typing'.
The rules are applied to all models.

\renewcommand{\bot}[1]{{\bf #1}}
\definecolor{LightCyan}{rgb}{0.88,1,1}
\begin{table*}[ht]
{\footnotesize
\setlength\tabcolsep{0.5ex}
\begin{tabular}{lL{2.5cm}L{2.5cm}L{2.5cm}}
\multicolumn{4}{l}{Friends of A}\\
\toprule
ID & Major & Company & Hobby \\
\midrule
1 & Metallurgical Engineering     & Gannett Company             & Candle making     \\
2 & Business Education & Electronic Arts & Gunsmithing \\
3 & Parks Administration & Kenworth & Water sports \\
\rowcolor{LightCyan}
4 & Mathematics Education & Electronic Arts &  Astronomy \\
5 & Agricultural Mechanization & AVST & Field hockey \\
6 & Mathematics Education & AVST & Shopping \\
7 & Parks Administration & Adobe Systems & Foreign language learning \\
8 & Agricultural Mechanization & Bronco Wine Company & Shopping \\
9 & Metallurgical Engineering & Electronic Arts & Foreign language learning \\
10 & Mathematics Education & Electronic Arts & Poi \\
\bottomrule
\end{tabular}
\quad
\begin{tabular}{lL{2.5cm}L{2.5cm}L{2.5cm}}
\multicolumn{4}{l}{Friends of B}\\
\toprule
ID & Major & Company & Hobby \\
\midrule
1 & Foreign Language Teacher Education & Gannett Company & Road biking \\
\rowcolor{LightCyan}
2 & Mathematics Education & Electronic Arts & Astronomy \\
3 & Petroleum Engineering & Western Sugar Cooperative & Candle making \\
4 & Mathematics Education & American Broadcasting Company & Road biking \\
5 & Petroleum Engineering & Western Sugar Cooperative & Road biking \\
6 & Petroleum Engineering & A\& W Restaurants & Golfing \\
7 & Petroleum Engineering & American Broadcasting Company & Origami \\
8 & Russian & The Walt Disney Company & Astronomy \\
9 & Petroleum Engineering & The Walt Disney Company & Origami \\
10 & Protestant Affiliation & Acme Brick & Astronomy \\
\bottomrule
\end{tabular}
\setlength\tabcolsep{0.5ex}
\begin{tabular}{L{7.5cm}L{7.5cm}}
\toprule
A: Human\quad B: Human & A: Human \quad B: \dkg{}\\
\midrule
\begin{tabular}[t]{L{7cm}}
A: hi \\
B: hi\\
A: Any friends work at AVST? \\
B: petroleum engi?? \eos{} no\\
A: no petroleum \\
B: russian as major? \\
A: Nope \\
B: protestant afiil? \eos{} math? \\
A: I have two math \\
B: SELECT 2 \\
A: one works at Electronic Arts \eos{} SELECT 10\\
B: SELECT 1 \\
A: SELECT 4 \quad B: SELECT 2\\
\end{tabular} &
\begin{tabular}[t]{L{7cm}}
B: \bot{hi}\\
A: hey\\
B: \bot{i have three friends who like road biking}\\
A: i have two friends who like foreign language learning\\
B: \bot{no foreign language learning here}   \\
A: I have no friends who like road biking\\
B: \bot{do they major in foreign language?} \\
A: No \eos{} no \\
B: \bot{petroleum engineering}? \\
A: No but I have four friends who work for the Electronic Arts company \\
B: \bot{SELECT 2} \quad A: SELECT 4 \\
\end{tabular} \\
\midrule
A: \skg{} \quad B: Human & A: Human \quad B: \rl{}\\
\midrule
\begin{tabular}[t]{L{7cm}}
A: \bot{hi \eos{} do you have any friends that work at electronic arts?} \\
B: One and they like astronomy \\
A: \bot{SELECT 10} \\
B: SELECT 2 \quad A: \bot{SELECT 4}\\

\end{tabular} &
\begin{tabular}[t]{L{7cm}}
B: \bot{hiya \eos{} i have one foreign language and gannett}\\
A: i have two avst\\
B: \bot{do you have any acme brick and astronomy?}\\
A: many maths people \eos{} two are foreign language\\
B: \bot{do you have any petroleum engineering and american broadcasting company?}\\
A: no \eos{} electronic arts\\
B: \bot{SELECT 1}\\
A: avst\\
B: \bot{do you have any disney or restaurant?} \\
...\\
\end{tabular} \\
\bottomrule
\end{tabular}
} 
\caption{\label{tab:human-bot-chats-2}
Example human-bot chats. The mutual friend is highlighted in blue in each KB.
Bots' utterances are in bold and
selected items are represented by item IDs.
Only the first half of the human-\rl{} chat is shown due to space limit. 
Multiple utterances of one agent is separated by $||$.
}
\end{table*}

\section{Additional Human-Bot Dialogue}
\label{sec:human-bot-chats}
We show another set of human-bot/human chats in \reftab{human-bot-chats-2}.
In this scenario, the distribution of values are more uniform compared to \reftab{human-bot-chats}.
Nevertheless, we see that \skg{} and \dkg{} still learned to start from relatively high-frequency entities.
They also appear more cooperative and mentions relevant entities in the dialogue context compared to \rl{}.

\section{Histograms of Ratings from Human Evaluations}
\label{sec:histograms}
\begin{figure*}[h]
\centering
\includegraphics[width=\textwidth]{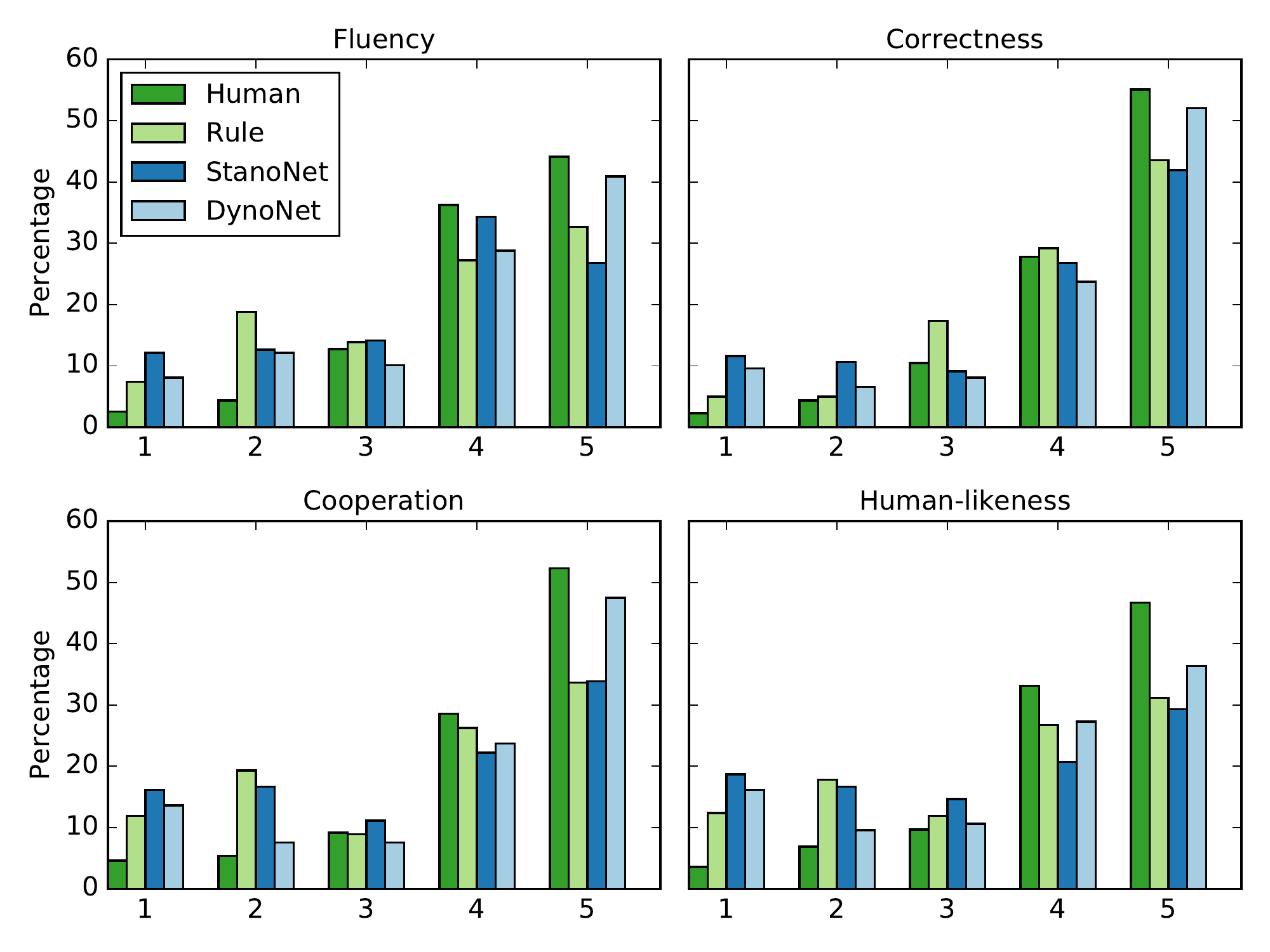}
\caption{\label{fig:partner-eval-dist} Histogram of ratings (higher is better) from dialogue partners.
\dkg{} is better than all other systems, especially in cooperation.
  }
\end{figure*}

\begin{figure*}[h]
\centering
\includegraphics[width=\textwidth]{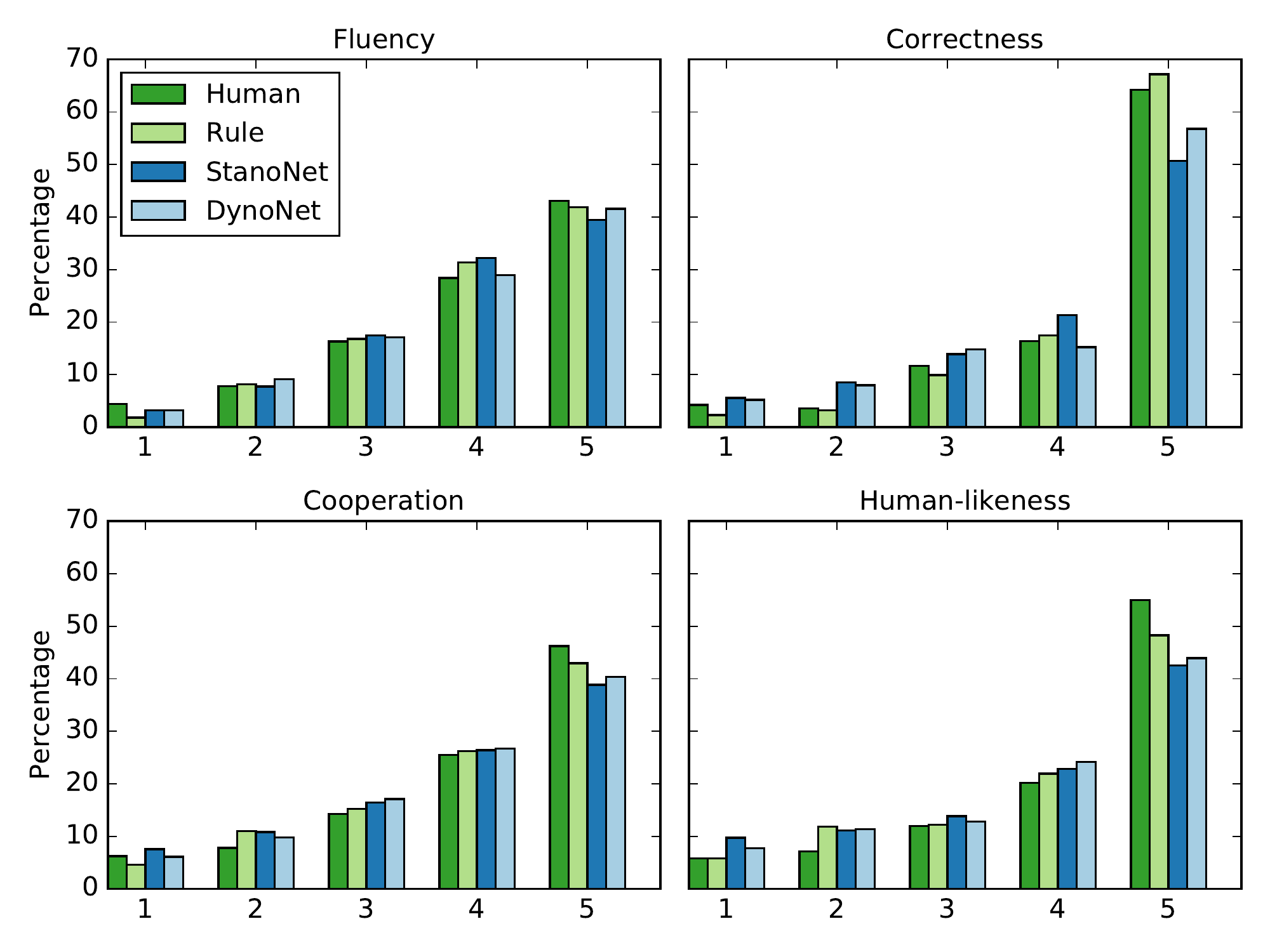}
\caption{\label{fig:third-eval-dist} Histogram of ratings (higher is better) from third-party evaluators. Differences between systems are less significant.}
\end{figure*}

The histograms of ratings from partner and third-party evaluations is shown in \reffig{partner-eval-dist} and \reffig{third-eval-dist} respectively.
As these figures show, there are some obvious discrepancies between the ratings made by agents who chatted with the bot and those made by an `objective' third party.
These ratings provide some interesting insights into how dialogue participants in this task setting perceive their partners, and what constitutes a `human-like' or a `fluent' partner.

\section{Example Comments from Partner and Third-party Evaluations}
\label{sec:comments}
In \reftab{comments}, we show several pairs of ratings and comments on human-likeness for the same dialogue from both the partner evaluation and the third-party evaluation.
As a conversation participant, the dialogue partner often judges from the cooperation and strategy perspective,
whereas the third-party evaluator relies more on linguistic features (e.g., length, spelling, formality). 

\begin{table*}
\footnotesize{
\centering
\begin{tabular}{ccL{3cm}|cc}
\toprule
\multirow{2}{*}{System} & \multicolumn{2}{c|}{Partner evaluation (1 per dialogue)} & \multicolumn{2}{c}{Third-party evaluation (5 per dialogue)} \\
& Human & \multicolumn{1}{c|}{Comments} & Human & Justifications \\
\midrule
Human & 4 & Good partner. Easy to work with & 4.6 &
\begin{tabular}[c]{@{}L{8cm}@{}}
- you have any friends who went to monmouth?\\
- The flow was nice and they were able to discern the correct answers. \\
- human like because of interaction talking\\
- Answers are human like, not robotic. Uses "hiya" to begin conversation, more of a warm tone.\\
- more human than computer Agent 2: hiya Agent 1: Hey
\end{tabular}  \\
\midrule
\rl{} & 2 & Didn't listen to me & 4 &
\begin{tabular}[c]{@{}L{8cm}@{}}
- agent 2 looked human to me\\
- definitely human\\
- A2 could be replaced with a robot without noticeable difference.\\
- They spoke and behaved as I or any human would in this situation.\\
- The agent just seems to be going through the motions, which gives me the idea that the agent doesn't exbit humanlike characteristics.
\end{tabular}  \\
\midrule
\skg{} & 5 & Took forever and didn't really respond correctly to questions. & 3.5 &
\begin{tabular}[c]{@{}L{8cm}@{}}
- No djarum -- This doesn't make sense in this context, so doesn't seem to be written by a human.\\
- human like because of slight mispellingss\\
- Can tell they are likely human but just not very verbose\\
- Their terse conversion leans to thinking they were either not paying attention or not human.\\
- The short vague sentences are very human like mistakes. 
\end{tabular}  \\
\midrule
\dkg{} & 4 & I replied twice that I only had indoor friends and was ignored. & 3.8 &
\begin{tabular}[c]{@{}L{8cm}@{}}
- Agent 1 is very human like based on the way they typed and the fact that they were being deceiving.\\
- Pretty responsive and logical progression, but it's very stilted sounding\\
- i donot have a jose\\
- Agent gives normal human responses, ``no angela i don't''\\
- agent 1 was looking like a humanlike
\end{tabular}  \\
\bottomrule
\end{tabular}
}
\caption{\label{tab:comments} Comparison of ratings and comments on human-likeness from partners and third-party evaluators.
Each row contains results for the same dialogue.
For the partner evaluation, we ask the human partner to provide a single, optional comment at the end of the conversation.
For the third-party evaluation, we ask five Turkers to rate each dialogue and report the mean score; they must provide justification for ratings in each aspect.
From the comments, we see that dialogue partners focus more on cooperation and effectiveness,
whereas third-party evaluators focus more on linguistic features such as verbosity and informality.
}
\end{table*}

\end{document}